\def\year{2018}\relax
\documentclass[letterpaper]{article} 
\usepackage{aaai18}  
\usepackage{times}  
\usepackage{helvet}  
\usepackage{courier}  
\usepackage{url}  
\usepackage{graphicx}  
\usepackage{amsmath}
\usepackage{amssymb}
\usepackage[flushleft]{threeparttable}
\usepackage{subcaption}
\DeclareMathOperator*{\argmax}{argmax}

\newcommand{\citet}[1]
{\citeauthor{#1}~\shortcite{#1}}
\newcommand{\citep}{\cite}

\title{A Multi-task Learning Approach for Improving Product Title Compression with User Search Log Data}
\author{Jingang Wang\textsuperscript{1}\thanks{Equal contributions.}, Junfeng Tian\textsuperscript{2}\footnotemark[1], Long Qiu\textsuperscript{3}, Sheng Li\textsuperscript{1}, Jun Lang\textsuperscript{1}, Luo Si\textsuperscript{1} \and Man Lan\textsuperscript{2,4} \\
	\textsuperscript{1} iDST, Alibaba Group  \\
	\textsuperscript{2} School of Computer Science and Software Engineering, East China Normal University  \\
	\textsuperscript{3} Onehome (Beijing) Network Technology Co. Ltd. \\
	\textsuperscript{4} Shanghai Key Laboratory of Multidimensional Information Processing, P.R.China \\
	\{jingang.wjg, lisheng.ls, langjun.lj, luo.si\}@alibaba-inc.com, \\
	qiulong@onehome.me, 51151201048@stu.ecnu.edu.cn, mlan@cs.ecnu.edu.cn}
\begin{document}
%

\maketitle
\begin{abstract}
It is a challenging and practical research problem to obtain effective compression of lengthy product titles for E-commerce. This is particularly important as more and more users browse mobile E-commerce apps and more merchants make the original product titles redundant and lengthy for Search Engine Optimization. Traditional text summarization approaches often require a large amount of preprocessing costs and do not capture the important issue of conversion rate in E-commerce. This paper proposes a novel multi-task learning approach for improving product title compression with user search log data. In particular, a pointer network-based sequence-to-sequence approach is utilized for title compression with an attentive mechanism as an extractive method and an attentive encoder-decoder approach is utilized for generating user search queries. The encoding parameters (i.e., semantic embedding of original titles) are shared among the two tasks and the attention distributions are jointly optimized. An extensive set of experiments with both human annotated data and online deployment demonstrate the advantage of the proposed research for both compression qualities and online business values.
\end{abstract}

\section{Introduction}
\noindent Mobile Internet are changing our lives profoundly. As depicted in the Statistical Report on China's Internet development, daily active users of mobile phones in China have overpassed 0.72 billion by June, 2017 \footnote{\url{http://www.cnnic.cn/hlwfzyj/hlwxzbg/hlwtjbg/201708/P020170807351923262153.pdf}}.
More online transactions are made on mobile phones instead of on PCs, with the gap still widening. This trend demands mobile E-Commerce platforms to improve user experience on their Apps. The most prominent distinction between smart phones and PCs lies in their screen sizes as a typical screen size of smart phones varies from 4.5 to 5.5 inches only.

\begin{figure}[htbp]
	\centering
	\begin{subfigure}[b]{0.225\textwidth}
	\includegraphics[width=\textwidth]{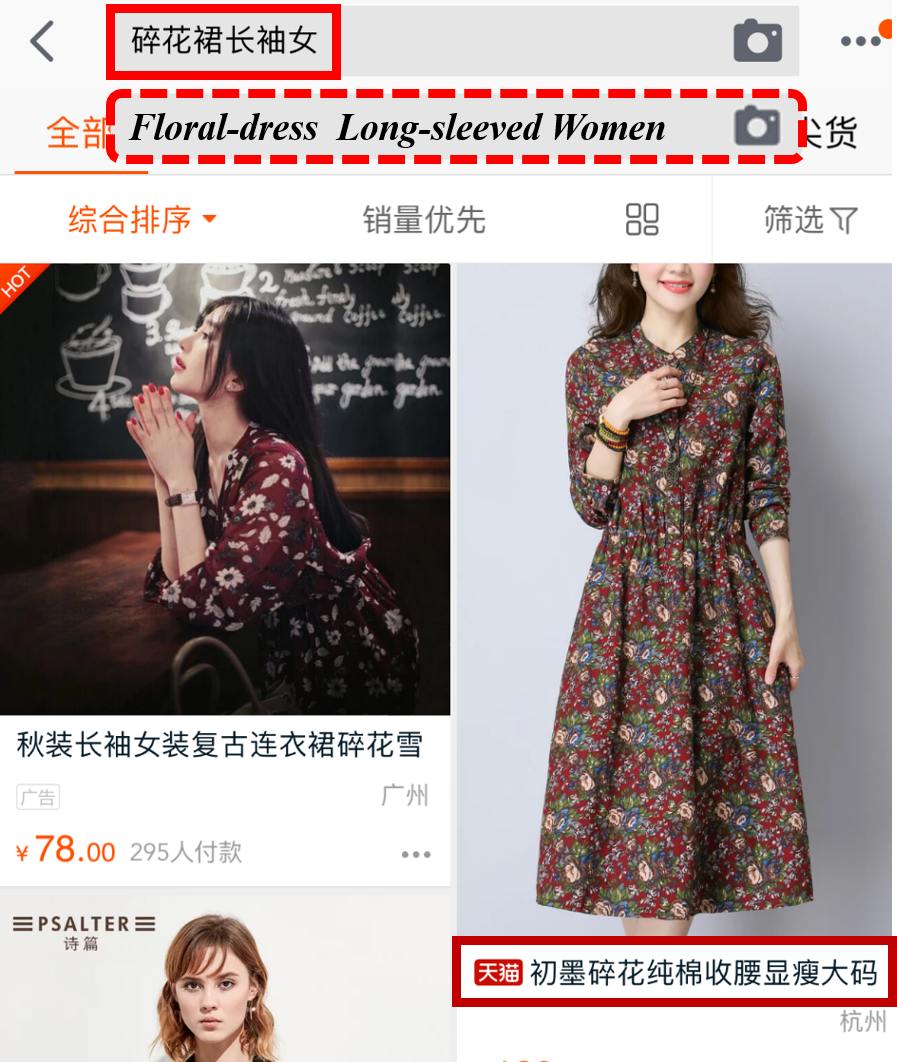}
	\caption{\emph{Search Result Page}}
	\label{subfig:srp}
	\end{subfigure}
	\begin{subfigure}[b]{0.225\textwidth}
	\includegraphics[width=\textwidth]{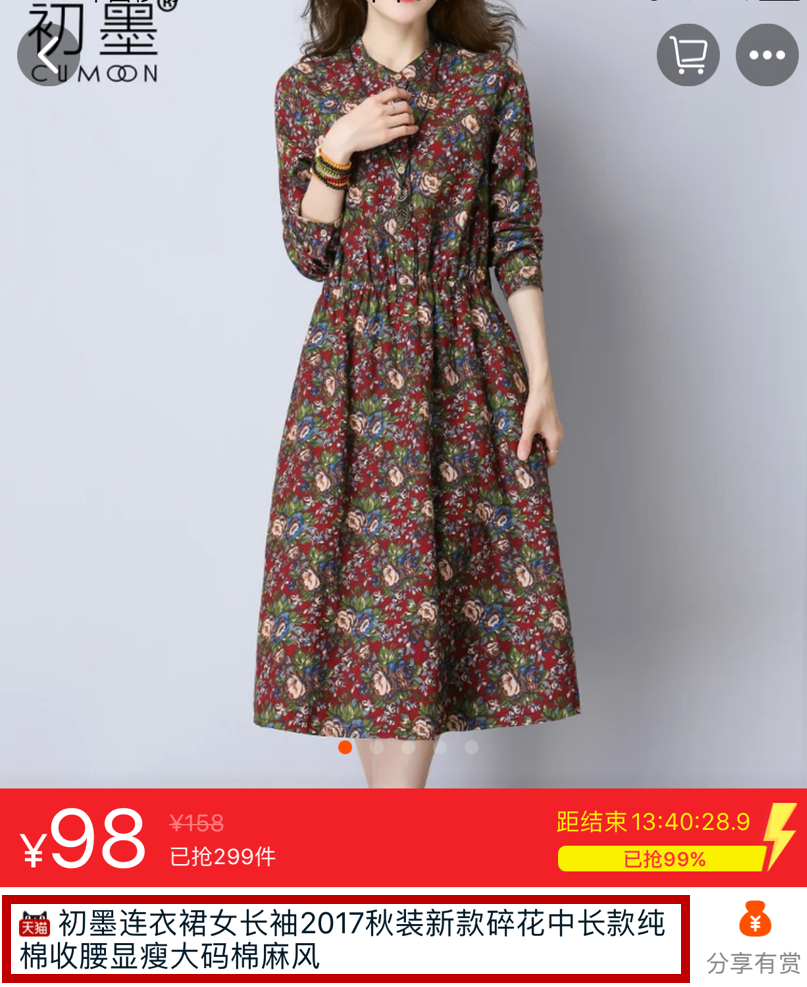}
	\caption{\emph{Product Detail Page}}
	\label{subfig:detail}
	\end{subfigure}
	\caption{When a user issues a query ``floral-dress long-sleeve women'', the complete title cannot be displayed in the Search Result Page, unless the user proceeds to the detail page further.}
	\label{fig:app}
\end{figure}

It is an important and practical research problem for producing succinct product titles in mobile E-commerce applications. On E-commerce platforms, especially customer to customer (C2C) websites, product titles are often written by the merchants. For the sake of SEO, most of these titles are rather redundant and lengthy. As shown in Figure \ref{subfig:detail}, the product title consists of more than 30 Chinese words, seriously hurting users' browsing experience. Usually, when a customer browses a product on Apps, less than 10 Chinese words could be displayed due to the screen size limit, as shown in Figure \ref{subfig:srp}. Amazon's research also reveals that product titles with less than 80 characters improve the shopping experience by making it easier for customers to find, review, and buy \footnote{\url{https://sellercentral.amazon.com/forums/message.jspa?messageID=2921001}}.

In comparison to conventional sentence summarization, which only requires the generated summary preserves important information grammatically, product title compression often has more constraints. The words in long titles are generally elaborated by experienced sellers and proved helpful to transaction in PC era. Many sellers do not want to include words not in their original titles. This is even more serious if the new external words make the purchase conversion rate lower. For this practical reason, we address product title compression as an extractive summarization task, i.e., all the words in the compressed short title are selected from the original product title.

Traditional methods for extractive title compression usually include two steps to perform: (1) word segmentation, fine-grained named entity recognition (NER), and term weighting preprocessing steps, and (2) constrained optimization (e.g., knapsack or integer linear programming (ILP) with pre-defined business rules and target length limit). The preprocessing steps are usually challenging and labor-intensive in E-commerce, because there exists hundreds of product categories (e.g., Clothes, Food and Electronics) on a typical E-commerce website and a single model usually performs sub-optimally across quite a lot of them. Recently emerging seq2seq models can generate sequential texts in a data-driven manner, getting rid of trivial labeling and feature engineering, but they require a large amount of human annotated training data. Furthermore, traditional summarization methods do not consider conversion rate, which is an important issue in E-commerce.

To address the problems of traditional summarization methods, this paper proposes a multi-task learning approach for improving product title compression with user search log data. For many E-commerce websites, both manually compressed titles and a large mount of user search log data exist. In particular, a multi-task framework is designed to include two attention-based neural networks. One network is used to model manually edited short titles and the original product tiles, and the other network models user search queries (with purchase transaction) with the original product titles. The two networks not only share the same encoder embedding (i.e., semantic embedding of original titles) but also are optimized simultaneously to reach an agreement on attention distributions over the original product title words. These properties enable the multi-task framework to generate more effective embedding and attention by making full use of manually edited titles, user queries and the transaction data. An extensive set of experimental results demonstrate that the multi-task framework not only achieves extractive title compression with a higher quality than alternatives but also improves conversion rates for more business values.

Although there are numerous sentence summarization research work, to the best of our knowledge, this is the first work focusing on product title compression in E-commerce. Our main contributions are as follows.
\begin{itemize}
	\item We propose a multi-task learning framework for extractive product title compression in E-commerce, outperforming traditional approaches like ILP.
	\item Our data-driven approach can save the labeling and feature engineering cost in conventional title compression approaches.
	\item The agreement-based setting on attention distributions is helpful to more accurately recognize important words in original titles, which is of great value in various E-commerce scenarios.
\end{itemize}

\section{Related Work}\label{sec:related-work}
Our work touches on several strands of research within text summarization, sentence compression and neural sequence modeling.
Existing text summarization methods can be categorized into extractive and abstractive methods.
Extractive methods produce a summary by dropping words from the original sentence, as opposed to abstractive compression which allows more flexible transformation.
Traditional extractive methods can be broadly classified into greedy approaches \citep{carbonell1998use,knight2000statistics}, graph-based approaches \citep{erkan2004lexrank}, and constraint optimization-based approaches \citep{mcdonald2007study,yao2017greedy}. Recently neural network based approaches have become popular for sentence summarization, especially abstractive summarization \citep{chopra2016abstractive,rush2015neural}.
In terms of extractive methods, \citet{cao2016attsum} propose AttSum to tackle extractive query-focused document summarization. The model learns query relevance and sentence saliency ranking jointly, and an attention mechanism is applied to select sentences when a query is given.
\citet{nallapati2017summarunner} present a recurrent neural network based sequential model for extractive document summarization, where each sentence is visited sequentially in the original order and a binary decision is made in terms of whether to preserve it in the summary.
\citet{cheng-lapata:2016:P16-1} develop a framework composed of a hierarchical document encoder and an attentive extractor for document/sentence extraction.
\citet{see2017get} propose Pointer-Generator network for summarization, which can copy words from the source text via pointing, achieving a balance between extractive and abstractive summarization.

More related to this work is sentence compression, in particular, compression by extraction.
\citet{mcdonald2006discriminative} employs linguistic features including deep syntactic ones to score the decision to extract each single word within its context, and decodes by dynamic programming to achieve an optimal extraction.
Similarly, \citet{thadani2013sentence} propose a constrained integer linear programming (ILP) framework to jointly consider n-grams and tree structures as extraction candidates.
More recently, \citet{filippova2015sentence} make word extraction decision based on Long Shot Term Memories (LSTMs).
\citet{miao2016language} adapt a variational auto-encoder (VAE) with a latent language model, which strives to keep the compression samples to closely resemble natural sentences.

Attention mechanism, which has gained popularity recently in multiple areas, allows the model to learn alignments between modularity \citep{luong2015effective,bahdanau2014neural,ba2014multiple,xu2015show}.
\citet{Cheng2016AgreementBasedJT} present agreement-based joint learning for bi-directional attentive neural machine translation, in which encoder-decoder components are trained with identical dataset in reverse directions.
In our work, two encoder-decoder components are trained with different data and agreement-based information is integrated into the loss objective.
\citet{luong2015multi} propose multi-task sequence-to-sequence learning (MTL) framework with different settings, including (1) the one-to-many setting sharing the encoder, (2) the many-to-one setting sharing the decoder, and (3) the many-to-many setting sharing multiple encoders and decoders.
Our model follows the one-to-many setting, where the encoder parameters are shared between short title generation and query generation.
In comparison, \citet{klerke2016improving}, dealing with a multi-task but not sequence-to-squence problem, optimize a shared Bi-LSTM network for extraction operation as well as gaze prediction.

\section{Data set}
Our proposed approach requires two parts of data, including (1) original product titles and their compressed versions, and (2) user search query and purchase log data for these products.
Since there doesn't exist an off-the-shelf benchmark data set for our task, we construct our data set from scratch.
We take advantage of realistic data from a well-known C2C website in China as our experimental data set.

In terms of the original title and compressed title pairs, we leverage the human-generated short titles from a product-recommendation channel of the website.
Display short titles in this channel are rewritten by professional business analysts in an extractive manner. We crawled all products belonging to the {\emph{Women's Clothes}} category as our experimental products.
We exclude the products whose original titles are shorter than 10 Chinese characters because product titles that are shorter than 10 characters can be displayed completely in most scenarios.

In terms of user search queries and purchase log data, we crawled user search queries leading to more than 10 transactions in one month given a product.

\begin{figure}[htbp]
	\centering
    \small
	\includegraphics[width=8.0cm]{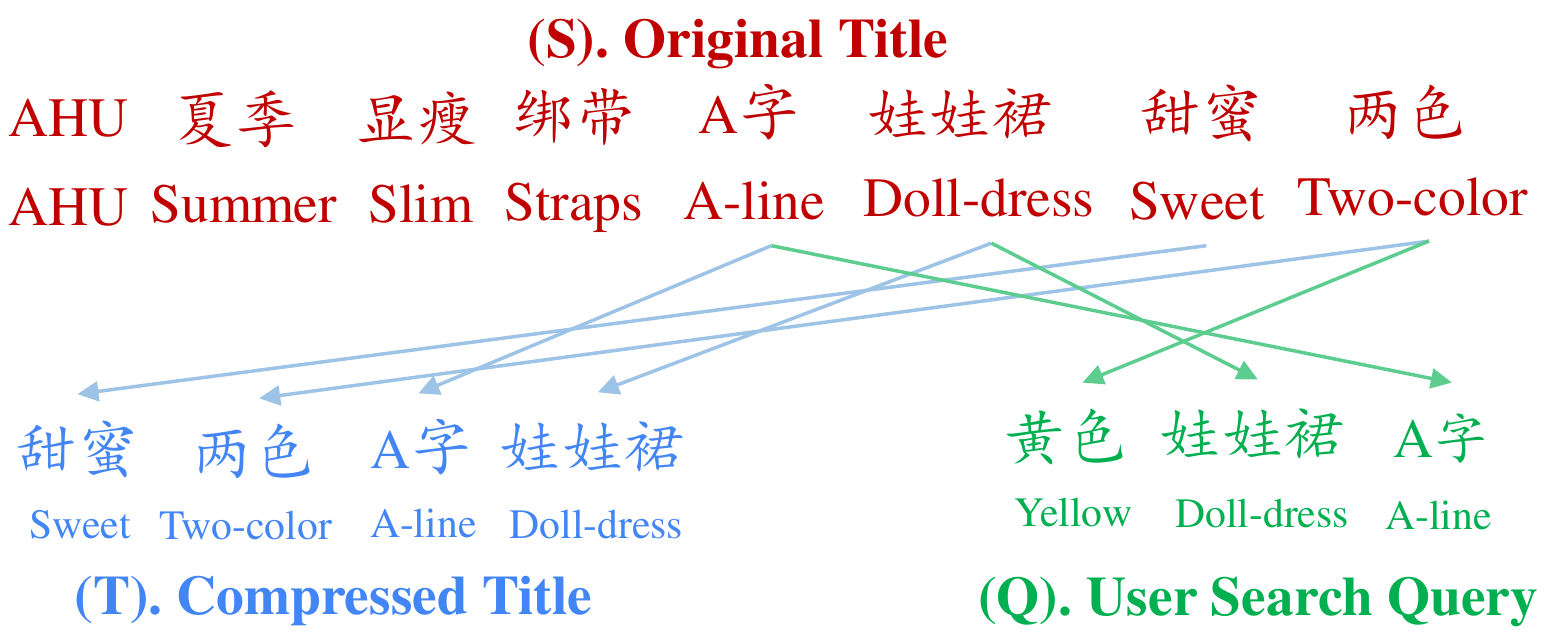}
	\caption{A triplet example. Both the compressed title and the query can help recognize important information from the original title.}
	\label{fig:example}
\end{figure}

The two parts of data are merged and organized as triplets, i.e., (product title, manually generated compressed title, user search query).
Therefore, the data set can be represented as $\langle S, T, Q \rangle$, where $S$ means products' original titles, $T$ means the handcrafted short titles, and $Q$ represents the successful transaction-leading search queries.
The details of our data set are shown in Table \ref{tb:dataset}, and a triplet example is presented in Figure \ref{fig:example}.
\begin{table}[htbp]
	\begin{threeparttable}
	\centering
	\caption{The statistics of the triplet data set}\label{tb:dataset}
	\begin{tabular}{l|c} \hline
		Data set (triplets) size  &  185, 386 \\
		Avg. length of original titles & 25.1\\
		Avg. length of handcrafted short titles & 7.5 \\
		Avg. length of search queries &  8.3 \\ \hline
	\end{tabular}
	\begin{tablenotes}
		\item Note: all the lengths are counted by Chinese characters, and each English word is counted as one Chinese character.
	\end{tablenotes}
	\end{threeparttable}
\end{table}

\section{Multi-task Learning for Product Title Compression}
\begin{figure*}[htbp]
	\centering
	\includegraphics[width=14cm]{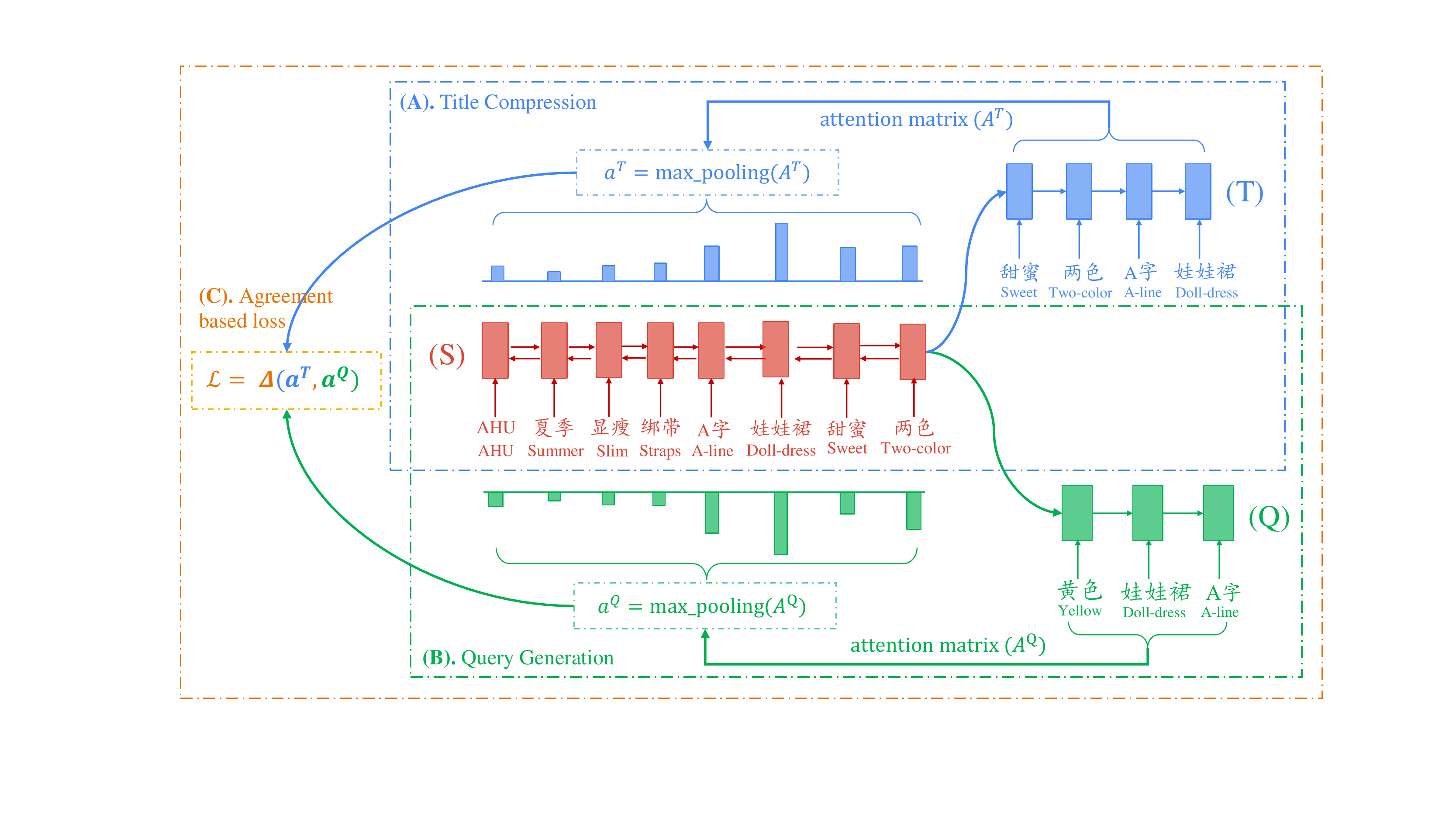}
	\caption{Multi-task Learning Framework, including two seq2seq components sharing the identical encoder. The main task is a Pointer Network to automatically point (select) the most informative words as compressed title. The auxiliary task is a standard seq2seq model to generate user search query. We utilize the attention distribution generated from user query to encourage the main task to agree on identity words.}
	\label{fig:mtl_model}
\end{figure*}

Given an input sequence, our goal is to produce a summary, where all the words are extracted from the input sequence. Suppose the input sequence $x$ contains $M$ words $x_{1}, x_{2},\ldots,x_{M}$ coming from a fixed vocabulary $\mathcal{V}$ of size $V$. A sequence to sequence (seq2seq) model takes $x$ as input and outputs a compressed sequence $y$ contains N words $y_{1}, y_{2},\ldots, y_{N}$ ($N < M$) coming from $\mathcal{V}$ as well. Assume the set $\mathcal{Y}$ as all possible sentences of length $N$, our objective is to find an optimal sequence from this set $\mathcal{Y}$. Please note that all the words $y_{i}$ in summary are transferred from the input sequence $x$ in this paper. Therefore, we have the objective function as

\begin{equation}\label{eq:obj}
\argmax_{y \in \mathcal{Y}}s(x, y) = \argmax_{m_i \in [1, M]} s(x, x_{[m_{1},\cdots, m_{N}]})
\end{equation}
where $s(x, y)$ is a scoring function. We represent it as the conditional log-probability of a summary given the input sequence, $s(x,y)=\log p(y|x;\theta)$, which can be rewritten as
\begin{equation}\label{eq:log-prob}
\log{p(y|x;\theta)} = \sum_{n=1}^{N}\log{P(y_{n}|x,y_{<n};\theta)}
\end{equation}
where $\theta$ is a set of model parameters and $y_{<n} = y_{1}, \ldots, y_{n-1}$ is a partial summary. The parameters $\theta$ of the model are learned by maximizing the Equation \ref{eq:log-prob} for the training set.

A basic seq2seq model consists of two recurrent neural networks (RNNs), named as encoder and decoder respectively \cite{kalchbrenner2013recurrent,cho-EtAl:2014:EMNLP2014,bahdanau2014neural}. The encoder processes the input sequence $x$ into a sequence of hidden states $h=h_{1},\ldots,h_{M}$,

\begin{equation}
h_{m} = f(x_{m}, h_{m-1}, \theta)
\end{equation}
where $h_{m}$ is the hidden state of the $m$-th input word and $f(\cdot)$ is a non-linear function. The final hidden state $h_{M}$ is the new initial state of the decoder. The decoder produces the hidden state $s=s_{1}, \ldots, s_{N}$ by receiving word embeddings of the previous words, and predicts next word according to the current hidden state.

In the basic encoder-decoder framework, the decoder is supposed to generate the output sequence solely based on the last hidden state $h_{M}$ from the encoder. It seems unreasonable to assume that the encoder can encode everything into a single state (a vector actually). RNNs are known to have problems dealing with such long-range dependencies. There are some tricks to tackle with this problem, such as replacing vanilla RNNs with Long Short Term Memories (LSTMs) or feeding an input sequence twice. However, the long-range dependencies are still problematic. Attention mechanism is introduced to address the problem \cite{bahdanau2014neural,luong2015effective}. With the attention mechanism, we no longer try to encode the full source sequence into a fixed-length vector. Rather, the decoder is allowed to ``attend'' to different parts of the source sequence at each step of the output generation. The conditional probability in Equation \ref{eq:log-prob} can be rewritten as
\begin{equation}
P(y_{n}|x,y_{<n};\theta) = g(y_{n-1},s_{n},c_{n}, \theta)
\end{equation}
where $g(\cdot)$ is a non-linear function, $s_{n}$ is the hidden state corresponding to the $n$-th target word computed by

\begin{equation}
s_{n} = f(s_{n-1},y_{n-1}, c_{n}, \theta)
\end{equation}
and $c_{n}$ is a context vector for generating the $n$-th target word,
\begin{equation}
c_{n} = \sum_{m=1}^{M}A(\theta)_{n,m}h_{m}
\end{equation}
where $A(\theta) \in \mathbb{R}^{N\times M}$ is referred as an attention matrix (i.e., alignment matrix in machine translation). $A(\theta)_{n,m}$ reflects the contribution of the $m$-th input word $x_{m}$ to generating the $n$-th word $y_{n}$.

\begin{equation}
A(\theta)_{n,m} = \frac{\exp(a(s_{n-1},h_{m},\theta))}{\sum_{m'=1}^{M}\exp(a(s_{n-1},h_{m'},\theta))}
\end{equation}
where $a(s_{n-1}, h_{m}, \theta)$ measures how well $x_{m}$ and $y_{n}$ are aligned, noted as $a_{n}^{m}$ in short,
\begin{equation}
a_{n}^{m} = v^{T}tanh(W_{1}s_{n-1} + W_{2}h_{m})
\end{equation}
where $v$, $W_{1}$, and $W_{2}$ are learnable parameters of the model.

Bearing the attentive seq2seq model in mind, we proceed to our multi-task framework containing two attentive seq2seq tasks for product title compression.
The structure of our multi-task framework is shown in Figure \ref{fig:mtl_model}.

The main task is fed with product titles and generates corresponding short titles, named as \emph{title compression}. Recall that the compressed titles are generated in an extractive manner. We implement the \emph{title compression} task with pointer networks \cite{vinyals2015pointer}, since the standard seq2seq model cannot be used for our compression problem where the output dictionary is composed of the words from the input sequence. Pointer networks do not blend the encode state $h_{m}$ to propagate extra information to the decoder, but instead, use $a_{n}^{m}$ as pointers to the input sequence directly.

The auxiliary task is fed with product titles and generates user search queries, named as \emph{query generation}.
We adopt the attentive encoder-decoder implementation for neural machine translation (NMT) \cite{bahdanau2014neural}.
The motivation lies that query generation can be deemed as a machine translation task seamlessly. Product titles are written by product-sellers (written in merchants' language), while search queries are issued by users (written in customers' language).
We believe the query generation task can contribute high-quality attention matrices on the original titles.

One intuitive approach for the multi-task setting is directly combine the losses of the two separate tasks, which cannot maximize the power of multi-task learning in our opinions.
Instead, we tie the two tasks in an elegant manner with agreement-based learning \cite{liang2008agreement,Cheng2016AgreementBasedJT}. Since our two tasks share an identical encoder which receives the original product title, we would obtain two attention distributions over the product title sequence.
We constrain these two distributions agree with each other, which means the important words recognized by two separate decoders are accordant.
With respect to implementation, we introduce an attention distribution agreement-based loss, as depicted in Equation \ref{eq:agree-loss}.
\begin{equation}\label{eq:agree-loss}
\mathcal{L}_{agree} = \mathcal{D}(A^{T}, A^{Q})
\end{equation}
$A^{T} \in \mathbb{R}^{N \times M}$ is the attention matrix of \emph{Title compression}, where $N$ and $M$ are the lengths of the generated short title and the original title respectively. $A^{Q} \in \mathbb{R}^{K \times M}$ is the attention matrix of \emph{Query generation}, where $K$ is the length of the user search query. Due to the different sizes of these two matrices, we perform max-pooling on them to get two vectors ($a^T, a^Q \in \mathbb{R}^{M}$) before calculating the agreement-based loss, as shown in Equation \ref{eq:max-pooling}.
\begin{equation}\label{eq:max-pooling}
\begin{split}
a^{T} = \max_{j=1}^{N} A_{j,:}^{T}; \quad a^{Q} = \max_{j=1}^{K} A_{j,:}^{Q}\\
\end{split}
\end{equation}
To evaluate the agreement between $a^{T}$ and $a^{Q}$, we adopt their KL-divergence as our agreement-based loss.
\begin{equation}\label{eq:KL}
	\mathcal{L}_{agree} =  KL \,(a^{T} \, \Vert \ a^{Q})
\end{equation}

Finally, the loss function of our multi-task framework becomes
\begin{equation}
\mathcal{L} = \lambda_{1}\mathcal{L}_{T} + \lambda_{2}\mathcal{L}_{Q} + (1-\lambda_{1}-\lambda_{2})\mathcal{L}_{agree}
\end{equation}
where $T$ represents \emph{Title compression} task, and \emph{Q} represents \emph{Query generation} task. $\lambda_{1}$ and $\lambda_{2}$ are hyper-parameters to tune the impacts of the two tasks.

\section{Experiments}
\subsection{Comparison Methods}
To evaluate the summarization performance of our approach, we implement rich extractive sentence summarization methods. Among them, the first two are common baselines.
\begin{itemize}
	\item Truncation-based Method (\textbf{Trunc.}). In most E-commerce scenarios where title compression is not available, long product titles are simply truncated to adapt to the limitation. Thus Truncation-based method is a na\"{\i}ve baseline for product title compression. Given a product title, we keep the words in their original order until the limit is reached.
	\item ILP-based Method (\textbf{ILP}). We also include the traditional ILP-based method \cite{clarke2008global} as another baseline, which is an unsupervised method that relies on preprocessing (i.e., word segmentation, NER and term weighting) results of the input titles.
	With respect to preprocessing, we adopt our internal Chinese processing toolkit.
	The term weighting algorithm is based on some heuristic rules. For example, the \emph{Product} words possess higher weights than \emph{Brand} words, and \emph{Brand} words possess higher weights than \emph{Modifier} words.
	Figure \ref{fig:ilp-example} presents an example of the preprocessing results given a product.
	Please note that the ILP-based approach is a rather strong baseline, which has been deployed in real scenarios of the mentioned E-commerce website.

\begin{figure*}[htbp]
	\centering
	\includegraphics[width=16cm]{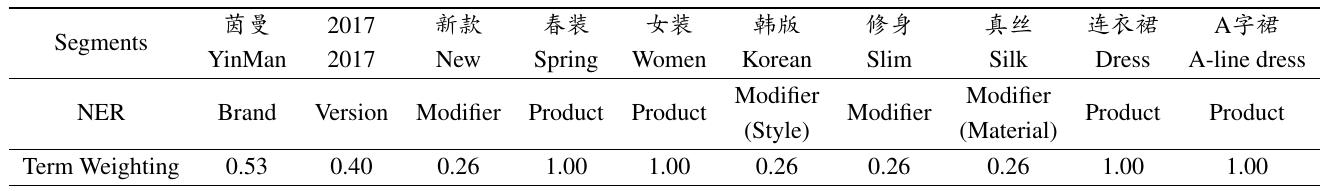}
	\caption{The  Chinese Word segmentation, NER and Term Weighting preprocessing results for a given product title.}
	\label{fig:ilp-example}
\end{figure*}
	\item Pointer Network (\textbf{Ptr-Net}). An attentive seq2seq model ``translating'' redundant long product titles to short ones. As we introduced before, to achieve the extractive summarization, we implement the Pointer Networks \cite{vinyals2015pointer}.
	\item \textbf{Vanilla MTL}. The proposed multi-task learning method, where the final loss is the linear combination of the two separate seq2seq tasks, i.e., $\mathcal{L} = \lambda\mathcal{L}_{T} + (1-\lambda)\mathcal{L}_{Q}$, the hyper-parameter $\lambda$ is tuned with the development data set. 
	\item Agreement-based MTL (\textbf{Agree-MTL}). The proposed attention distribution agreement-based multi-task learning approach. In implementation, the attention distribution agreement-based loss is interpolated into the loss function during model training, i.e., $\mathcal{L} = \lambda_{1}\mathcal{L}_{T} + \lambda_{2}\mathcal{L}_{Q} + (1-\lambda_{1}-\lambda_{2}) \mathcal{L}_{agree}$. The hyper-parameters are tuned with the development data set.
\end{itemize}

\subsection{Seq2seq Model Settings and Parameter Tuning}\label{subsec:tuning}
For the implementation of our seq2seq models (i.e., Ptr-Net, Vanilla-MTL and Agree-MTL), we adopt two 128-dimensional LSTMs for the bidirectional encoder and one 256-dimensional LSTMs for the decoder.
To avoid the effect of Chinese word segmentation error, both our encoder and decoder inputs are Chinese characters instead of words.
We initialize a 128-dimensional word embedding following normal distribution  $\mathcal{N}\left(0, 1\mathrm{e}{-4}^{2}\right)$. All the models are trained on a single Tesla M40 GPU, and optimized with Adagrad \cite{duchi2011adaptive} (learning rate=$0.15$, and batch size=$128$). We use gradient clipping with a maximum gradient norm of $2$, but do not use any form of regularization.
At test time, our short titles are produced with a decoder whose beam search size is 10 and maximum decoding step size is 12.
We randomly select 80\% of the triplet data as training data, and remain the remainder for development and test (10\% for each).
We first tune $\lambda$ of Vanilla MTL using grid search on [0.1, 0.9].
The model performance cannot be improved further when $\lambda \geq 0.5$. Therefore $\lambda$ of Vanilla MTL is set as $0.5$.
Then we conduct the parameter tuning for Agree-MTL. We fix $\lambda_1$ as $0.5$ and further tune $\lambda_2$.
The model achieves best over the development data set when $\lambda_2$ reaches $0.3$.

\subsection{Automatic Evaluation}
Summarization systems are usually evaluated using several variants of the recall-oriented ROUGE metric \cite{lin2004rouge}.
ROUGE measures the summary quality by counting the overlapping units such as n-grams between the generated summary and reference summaries.
We consider ROUGE-1 (uni-grams), ROUGE-2 (bi-grams) and ROUGE-L (longest-common subsequence) as our automatic evaluation metrics. The results are shown in Table \ref{tb:rouge-result}.
As expected, the truncation-based method achieves the worst ROUGE scores because of its ignorance of semantics.
The main reason is that core product words are usually appear in the latter part of the titles, and truncation-based methods tend to miss some of them in the compressed titles by taking only the lead.

Although a reversed variant of the truncation-based method may mitigate the problem, it risks missing some important brand names that normally appear early in the titles.
Compared with truncation-based method, ILP-method can improve the ROUGE-1 by 18 percents and the ROUGE-2 by 10 percents. This gap may be resulted from Chinese word segmentation mistakes.
All the three seq2seq variants perform better than ILP-based method obviously, revealing that seq2seq models are more capable of imitating edited short titles than unsupervised methods.
The Vanilla-MTL without any constraint on two separate seq2seq tasks cannot beat the independent seq2seq task (i.e., Ptr-Net).
Nevertheless, the introduction of attention distribution agreement-based constraint can enhance the performance obviously, and perform best on all variants of ROUGE metrics.

\begin{table}[htbp]
\begin{threeparttable}
	\centering
	\caption{ROUGE performance of various methods on the test set.}\label{tb:rouge-result}
	\begin{tabular}{lccc} \hline
		Method  &  ROUGE-1 & ROUGE-2 & ROUGE-L \\
		\hline
		Trunc.       &  30.43    & 19.13   & 29.00 \\
		ILP          &  48.28    & 29.84   & 43.65 \\
		\hline
		Ptr-Net      &  69.03    & 55.30   & 67.98 \\
		\hline
		Vanilla-MTL  &  65.92    & 52.94   & 65.20 \\
		Agree-MTL & \textbf{70.89} & \textbf{56.80} & \textbf{69.61} \\ \hline
	\end{tabular}
\end{threeparttable}
\end{table}

\subsection{Manual Evaluation}
Like related summarization work \cite{filippova2013overcoming,tan2017abstractive}, we also conduct manual evaluation on the generated short titles.
We randomly sampled 300 products in the Women's Clothes category, and asked three participants to annotate the quality of generated short titles.
Three perspectives are considered during manual evaluation process: (1) \textbf{Core Product Recognition}. Is the core product word detected correctly in the compressed title? (2) \textbf{Readability}. How fluent, grammatical the short title is? (3) \textbf{Informativeness}. How informative the short title is?

Compared with other sentence summarization work, core product recognition is a particular requirement for product title compression.
Consider a product with a Chinese title ``2017 new-style Women's Dress Slim Bowknot One-Piece Skirt'', there exists three product words in the original title, including ``Women's Dress'', ``bowknot'' and ``one-piece dress''.
Obviously, ``Women's Dress'' is too general to be a good core product term, and ``bowknot'' is a noun-modifier of ``one-piece dress''. A good compression method should reserve the core product term and drop the other redundant ones.

The \textbf{core product recognition} property is assessed with a binary score (i.e, $1$ for correct and $0$ for incorrect), and the other two properties are assessed with a score from $1$ (worst) to $5$ (best).
The results are presented in Table \ref{tb:human-result}. To make annotation results more consistent and reliable, we exclude instances with divergent ratings (i.e., their variance is larger than a threshold). We conducted paired t-tests on the manual evaluation results. With respect to readability, our Agree-MTL is significantly (p $<$ 0.05) better than all other methods. With respect to informativeness, our Agree-MTL is significantly (p $<$ 0.05) better than all other methods except Ptr-Net (it would be significantly better than Ptr-Net when p $<$ 0.08)

The truncation-based method only achieves a accuracy of 8.3\% on core product recognition.
As we analyzed in last section, this is caused by the phenomenon that core product words usually appear in the latter part of product titles.

The results indicate that our MTL methods outperform the unsupervised methods (i.e., \textbf{Trunc.} and \textbf{ILP}), getting particularly good marks for both readability and informativeness.
Note that the unsupervised baseline is also capable of generating readable compressions but does a much poorer job in selecting most important information. Our MTL method successfully learned to optimize both scores.

\begin{table}[htbp]
  \begin{threeparttable}
	\centering
	\caption{Manual evaluation results, including average core product recognition accuracy (Avg. Accu), average readability score (Avg. Read) and average informativeness score (Avg. Info).
    }\label{tb:human-result}
	\begin{tabular}{lccc} \hline
		Method  &  Avg. Accu & Avg. Read & Avg. Info  \\ \hline
		Trunc.  & 8.33 \% & 1.93 & 1.96 \\
		ILP     & 93.33\% & 4.63 & 3.90 \\ \hline
		Ptr-Net & \bf 98.33 \% & 4.66 & 4.13 \\
		\hline
		Vanilla-MTL &  96.67\% & 4.63 & 3.90 \\
		Agree-MTL   &  \bf 98.33 \% & \bf 4.80 & \bf 4.66\\ \hline
	\end{tabular}
  \end{threeparttable}
\end{table}

\subsection{Case Studies}
\begin{figure*}[htbp]
	\centering
	\includegraphics[width=16cm]{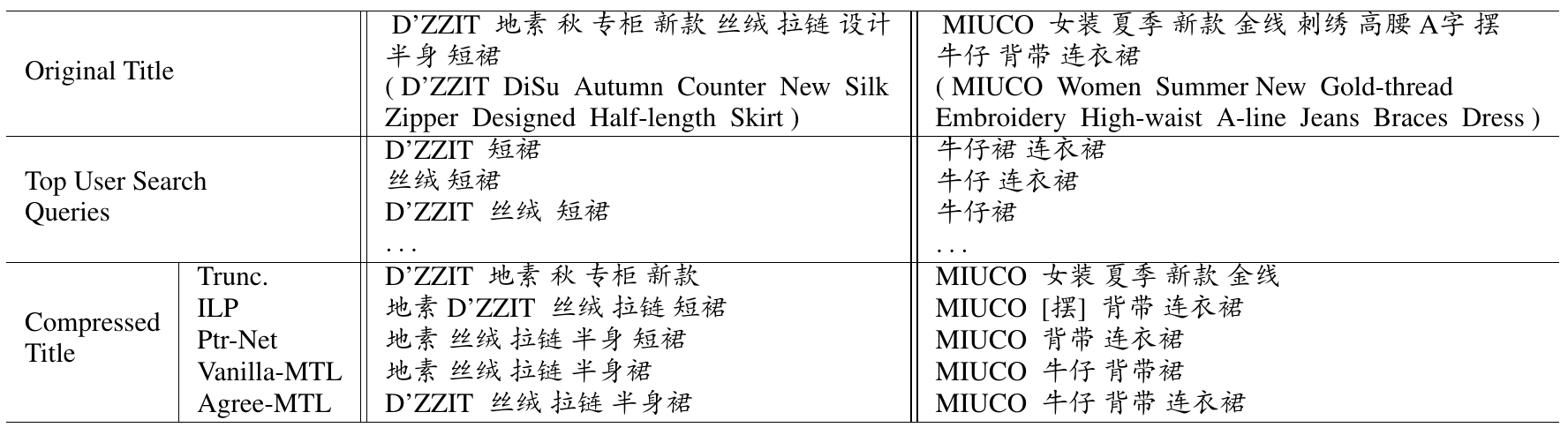}
	\caption{Case studies.}\label{fig:case-study}
\end{figure*}

In this section, we present two realistic cases in Figure \ref{fig:case-study}.

Generally, truncation-based method achieves the worst performance because of missing core product words in the short titles.
ILP-based method can recognize core product words correctly and produce reasonable short titles. However, readability is sometimes unsatisfactory. In the right example of Figure \ref{fig:case-study}, the short title generate by ILP is not fluent due to word segmentation mistakes (surrounded by square brackets).

In comparison to both baselines, three seq2seq-based methods can produce fluent and grammatic short titles.
With the presence of attention distribution agreement constraint, Agree-MTL perform better than Vanilla-MTL and Ptr-Net by exposing important words that match users' information need (i.e., top search queries) on the corresponding product. Consider the left example of Figure \ref{fig:case-study}, Agree-MTL method produces the short title with the English brand word (i.e., \emph{D'ZZIT}), while Vanilla-MTL uses the Chinese brand word (i.e., \emph{DiSu}) instead. From the top search queries that successfully leads final transactions, we can conclude that users prefer to search this product with the English brand word, and Agree-MTL captures this information and exposes it during compression.

\section{Online Deployment}
Previous experimental results have proven the advantages of our proposed Agree-MTL approach, so we deploy it in a real world online environment to test its practical performance.

We perform A/B testing in the search result page scenario of the mentioned E-commerce website (over 4 million daily active users in the Women's Clothes category).
Note that ILP-based method is already deployed on the website.
The A/B testing system split online users equally into two groups and direct them into two separate buckets respectively.
Then for users in the A bucket (i.e., the baseline group), the short titles are generated by ILP-based method. While for users in the B bucket (i.e., the experimental group), the short titles are generated by Agree-MTL method.

The online A/B testing lasted for one week.
All the conditions of the two buckets are identical except the title compression methods.
Two indicative measures are adopted to test the performance, including product Click Through Rate (CTR) and Click Conversion Rate (CVR) which can be calculated as
\begin{equation}\label{eq:ctr-cvr}
\begin{split}
CTR = \frac{\#product\_click}{\#product\_PV} \\
CVR = \frac{\#product\_trade}{\#product\_click}
\end{split}
\end{equation}
where $\#product\_click$ is the clicking times of the product, $\#product\_PV$ is the page views of the product, and $\#product\_trade$ is the number of purchases of the product.

We calculated the overall CTR and CVR for all the products in the Women's Clothes category in the two buckets.
We found that the performance in the experimental bucket (i.e., Agree-MTL) is significantly better ($p < 0.05$) than that in the baseline bucket (i.e., ILP) on both measures.
Agree-MTL improved the CTR by 2.58\% and the CVR by 1.32\% over the baseline.
More specifically, the improvement of CTR implies that users are more likely to browser and click the Agree-MTL generated short titles.
The improvement of CVR means that after a user's click, Agree-MTL has higher probability to convert the click into a purchase action.
This is quite impressive, if we consider the fact that product titles on a search result page occupy a relatively small space and thus only partially affect the users' decision.

\section{Conclusion}
Product titles on E-commerce platforms are frequently redundant and lengthy as a result of SEO.
Meanwhile, people are getting used to browsing E-commerce Apps on their phones, where long titles cannot be displayed properly due to the limited screen size. Hence, product title compression is much desired.
Taking it as an extractive sentence summarization task, traditional methods require lots of preprocessing cost without taking transaction conversion rate optimization into consideration.
We address this problem with a pointer network-based sequence to sequence model in a multi-task setting. Sharing an identical encoder, a parallel seq2seq model is trained to match product titles and their transaction-leading search queries. We apply the attention mechanism in this framework, where for the two seq2seq models their individual attention distributions over tokens in product titles are jointly optimized to better focus on important tokens during compression.
This framework not only improves user experience by compressing redundant titles into concise ones, but also guarantees query-initiated transaction conversion rate by prioritizing query-related keywords in the resultant short titles.
We perform extensive experiments with realistic data from a dominant Chinese E-commerce website.
The advantages of the proposed framework in terms of its compression quality and business value are demonstrated by these experiments and online deployment.


\section{Acknowledgments}
The authors would like to thank Dr. Nan Li and Dr. Ding Liu with Alibaba Group for their valuable discussions and the anonymous reviewers for their helpful comments.
Junfeng Tian is supported by the Science and Technology Commission of Shanghai Municipality
Grant (No. 15ZR1410700) and the Open Project of Shanghai Key Laboratory of Trustworthy Computing Grant (No. 07dz22304201604).

\bibliographystyle{aaai}
\bibliography{Wang-Tian}

\end{document}